\documentclass[journal,10pt,compsoc]{IEEEtran}

\ifCLASSINFOpdf
\else
\fi
\usepackage{epsfig,float}
\usepackage{picinpar}
\usepackage{graphicx}
\usepackage{epstopdf}
\usepackage{amsmath}
\usepackage{amsthm}
\usepackage{amssymb,bm}
\usepackage[ruled,vlined]{algorithm2e}
\usepackage{float,caption,multirow}
\usepackage{url,caption}
\usepackage{ragged2e}
\usepackage{subcaption}
\usepackage{color}

\newcommand{\defeq}{\mathrel{\mathop:}=}

\newcommand{\bs}{\mathbf}
\newcommand{\mc}{\mathcal}

\begin{document}

\title{Kindling the Darkness:\\ A Practical Low-light Image Enhancer}

\author{Yonghua~Zhang,  Jiawan  Zhang,  and Xiaojie~Guo
% <-this % stops a space
%\thanks{Manuscript received Aug. 30, 2017; revised Jun. 15, 2018 and Nov.20, 2018; X. Guo was supported by NSFC (grant no. 61772512) and CCF-Tencent Open Research Fund. J. Ma was supported by NSFC (grand no. 61773295). H. Ling was supported by XXX.}

\thanks{$\bullet$ Y. Zhang (zhangyonghua@tju.edu.cn), J. Zhang (jwzhang@tju.edu.cn), and X. Guo (xj.max.guo@gmail.com) are with the College of Intelligence and Computing,
Tianjin University, Tianjin 300350, China.}
\thanks{Corresponding author:  X. Guo (xj.max.guo@gmail.com)}
}

%}% <-this % stops a space

% The paper headers
\markboth{}%IEEE Transactions on Pattern Analysis and Machine Intelligence} %
{Guo \textit{et al.}:  Mutually Guided Image Filtering}

%\maketitle

\IEEEtitleabstractindextext{

\begin{abstract}
Images captured under low-light conditions often suffer from (partially) poor visibility. Besides unsatisfactory lightings, multiple types of degradations, such as noise and color distortion due to the limited quality of cameras, hide in the dark. In other words, solely turning up the brightness of dark regions will inevitably amplify hidden artifacts. This work builds a simple yet effective network for \textbf{Kin}dling the \textbf{D}arkness (denoted as KinD), which, inspired by Retinex theory, decomposes images into two components. One component (illumination) is responsible for light adjustment, while the other (reflectance) for degradation removal. In such a way, the original space is decoupled into two smaller subspaces, expecting to be better regularized/learned. It is worth to note that our network is trained with paired images shot under different exposure conditions, instead of using any ground-truth reflectance and illumination information. Extensive experiments are conducted to demonstrate the efficacy of our design and its superiority over state-of-the-art alternatives. Our KinD is robust against severe visual defects, and user-friendly to arbitrarily adjust light levels. In addition, our model spends less than 50ms to process an image in VGA resolution on a 2080Ti GPU. All the above merits make our KinD attractive for practical use. 
\end{abstract}

}

\IEEEpeerreviewmaketitle

\maketitle

\section{Introduction}
\IEEEPARstart{V}{ery} often, capturing high-quality images in dim light conditions is challenging. Though a few operations, such as setting high ISO, long exposure, and flash, can be applied under the circumstances, they suffer from different drawbacks.  For instance, high ISO increases the sensitivity of an image sensor to light, but the noise is also amplified, thus leading to the low (signal-to-noise ratio) SNR. Long exposure is limited to shoot static scenes, otherwise it highly likely gets in trouble of blurry results. Using flash can somehow brighten the environment, which however frequently introduces unexpected highlights and unbalanced lighting into photos, making them visually unpleasant. In practice, typical users may even not have the above options with limited photographing tools, {\it e.g.} cameras embedded in portable devices. Although the low-light image enhancement has been a long-standing problem in the community with a great progress made over the past years, {\it developing a practical low-light image enhancer remains challenging, since flexibly lightening the darkness, effectively removing the degradations, and being efficient should all be concerned.} % To obtain satisfactory images under low light conditions, low-light image enhancement techniques are definitely required.

%\begin{figure*}[t]
%	\begin{center}
%		\begin{subfigure}{0.32\linewidth}
%			\includegraphics[width=1\linewidth]{fig/3.png}
%		\end{subfigure}
%	\begin{subfigure}{0.32\linewidth}
%		\includegraphics[width=1\linewidth]{fig/8input.png}
%	\end{subfigure}
%	\begin{subfigure}{0.32\linewidth}
%	\includegraphics[width=1\linewidth]{fig/1input.png}
%\end{subfigure}\\
%		\begin{subfigure}{0.32\linewidth}
%			\includegraphics[width=1\linewidth]{fig/3output.png}
%		\end{subfigure}
%		\begin{subfigure}{0.32\linewidth}
%			\includegraphics[width=1\linewidth]{fig/Balloons_kindle.png}
%		\end{subfigure}
%		\begin{subfigure}{0.32\linewidth}
%			\includegraphics[width=1\linewidth]{fig/1output.png}
%		\end{subfigure}
%	\end{center}
%	\caption{Left column: three natural images captured under different light conditions. Right column: our enhanced results. Notice that the first image is with extremely low light, we show its x20 version on the top-right corner.}
%	\vspace{-10pt}
%	\label{fig:open}
%\end{figure*}

\begin{figure}[t]
	\begin{center}
			\includegraphics[width=1\linewidth]{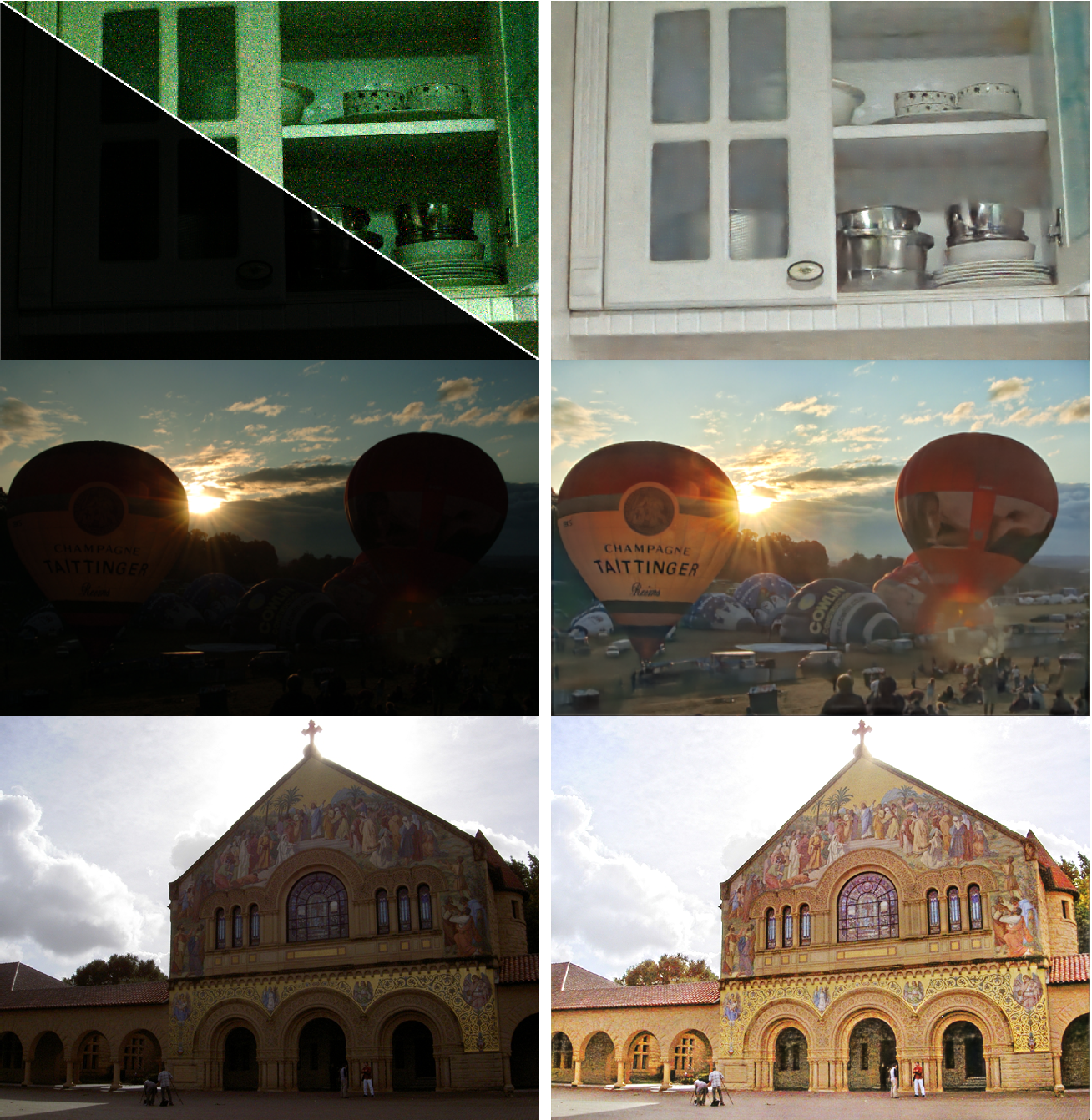}
	\end{center}
	\caption{Left column: three natural images captured under different light conditions. Right column: our enhanced results. Notice that the first image is with extremely low light, we show its x20 version on the top-right corner.}
	\vspace{-10pt}
	\label{fig:open}
\end{figure}

Figure \ref{fig:open} provides three natural images captured under challenging light conditions. Concretely, the first case is with extremely low light. Severe noise and color distortion are hidden in the dark. By simply amplifying the intensity of the image, the degradations show up as given on the top-right corner. The second image is photographed at sunset (weak ambient light), most objects in which suffer from backlighting.  Imaging at noon facing to the light source (the sun) also hardly gets rid of the issue like the second case exhibits, although the ambient light is stronger and the scene is more visible. Note that those relatively bright regions of the last two photos will be saturated by direct amplification.

Deep learning-based methods have revealed their superior performance in numerical low-level vision tasks, such as denoising and super-resolution, most of which need the training data with ground truth. For the target problem, say low-light image enhancement, {\it no ground-truth real data exists}, although the order of light intensity can be determined. Because, from the viewpoint of users, the favorite light levels for different people/requirements could be much diverse. In other words, one cannot say what light condition is the best/ground-truth. Therefore, it is not so felicitous to map an image only to a version with a specific level of light.

Based on the above analysis, we summarize challenges in low-light image enhancement as follows:
\emph{\begin{itemize}
		\item How to effectively estimate the illumination component from a single image, and flexibly adjust light levels?
		\item How to remove the degradations like noise and color distortion previously hidden in the darkness after lightening up dark regions?
		\item How to train a model without well-defined ground-truth light conditions for low-light image enhancement by only looking at two/several different examples?
\end{itemize}} 
\noindent In this paper, we propose a deep neural network to take the above concerns into account simultaneously.

\subsection{Previous Arts}
\label{sec:PA}
A large number of low-light image enhancement schemes have been proposed. In what follows, we briefly review classic and contemporary works closely related to ours. % In the literature

\textbf{Plain Methods.} Intuitively, for an image with the globally low light, the visibility can be enhanced by directly amplifying it. But, as shown in the first case of Figure \ref{fig:open},  the visual defects including noise and color distortion show up along the details. For images containing bright regions, {\it e.g.} the last two pictures in Figure \ref{fig:open}, this operation easily results in (partial) saturation/over-exposure. One technical line, with histogram equalization (HE) \cite{Pisano1998Contrast,Cheng2004A,DHEJ} and its follow-ups \cite{Turgay2011Contextual,Chulwoo2013Contrast} as representatives, tries to map the value range into [0, 1] and balance the histogram of outputs for avoiding the truncation problem. These methods {\it de facto} aim to increase the contrast of image. Another mapping manner is gamma correction (GC), which is carried out on each pixel individually in a non-linear way. Although GC can promote the brightness especially of dark pixels, it does not consider the relationship of a certain pixel with its neighbors. {\it The main drawback of the plain approaches is that they barely consider real illumination factors, usually making enhanced results visually vulnerable and inconsistent with real scenes.}

\textbf{Traditional Illumination-based Methods.} Different from the plain methods, strategies in this category are aware of the concept of illumination. The key assumption, inspired by Retinex theory \cite{Land1977The}, is that the (color) image can be decomposed into two components, {\it i.e.} reflectance and illumination. Early attempts include single-scale Retinex (SSR) \cite{SSR} and multi-scale Retinex (MSR) \cite{MSR}. Limited to the manner of producing the final result, the output often looks unnatural and somewhere over-enhanced. Wang {\it et al.} proposed a method called NPE \cite{NPE}, which jointly enhances contrast and preserves naturalness of illumination. Fu {\it et al.} developed a method \cite{Fu2016A}, which adjusts the illumination through fusing multiple derivations of the initially estimated illumination map. However, this method sometimes sacrifices the realism of those regions containing rich textures. Guo {\it et al.} focused on estimating the structured illumination map from an initial one \cite{LIME}. {\it  These methods generally assume that the images are noise- and color distortion-free, and do not explicitly consider the degradations.} In \cite{SRIE}, a weighted variational model for simultaneous reflectance and illumination estimation (SRIE) was designed to obtain better reflectance and illumination layers, then the target image is generated by manipulating the illumination. Following \cite{LIME}, Li {\it et al.} further introduced an extra term to host noise \cite{RRM}. {\it Although both \cite{SRIE} and \cite{RRM} can reject slight noise in images, they are short of abilities in handling color distortion and heavy noise.} 

\textbf{Deep Learning-based Methods.} With the emergence of deep learning, a number of low-level vision tasks have been benefited from deep models, such as \cite{NIPSDen,DnCNN} for denoising, \cite{PAMISR} for super-resolution, \cite{ICCVCmp} for compression artifact removal and \cite{DehNet} for dehazing. Regarding the target mission of this paper,  the low-light net (LLNet) proposed in \cite{LLNet} builds a deep network that performs as a simultaneous contrast enhancement and denoising module. Shen {\it et al.}  deemed that multi-scale Retinex is equivalent to a feed-forward convolutional neural network with different Gaussian convolution kernels. Motivated by this, they constructed a convolutional neural network (MSR-net) \cite{MSR-net} to learn an end-to-end mapping between dark and bright images. Wei {\it et al.} designed a deep network, called Retinex-Net \cite{DRD}, that integrates image decomposition and illumination mapping. Please notice that Retinex-Net additionally employs an off-the-shelf denoising tool (BM3D \cite{BM3D}) to clean the reflectance component. {\it These strategies all assume that there exist images with ``ground-truth" lights, without considering that the noise differently affects regions with various lights.} Simply speaking, after extracting the illumination factor, the noise level of dark regions is (much) higher than that of bright ones in the reflectance. In such a situation, adopting/training a denoiser with a uniform ability over an image (reflectance) is no longer suitable. In addition, the above methods do not explicitly cope with the degradation of color distortion, which is not uncommon in real images. More recently, Chen {\it et al.} proposed a pipeline for processing low-light images based on end-to-end training of a fully convolutional network \cite{SID}, which can jointly deal with noise and color distortion. However, this work is specific to data in RAW format, limiting its applicable scenarios. As stated in \cite{SID}, if modifying the network to accept data in JPEG format, the performance significantly drops.

Most existing methods manipulate the illumination by gamma correction, appointing a level existing in carefully constructed training data, or fusion. For gamma correction, it may be unable to reflect the relationship between different light (exposure) levels. As for the second manner, it is heavily restricted to whether the appointed level is contained in the training data. While for the last one, it even does not provide a manipulation option. Therefore, {\it it is desired to learn a mapping function to arbitrarily convert one light (exposure) level to another for offering users the flexibility of adjustment.}

\begin{figure*}[t]
	\begin{center}
		\includegraphics[width=1\linewidth]{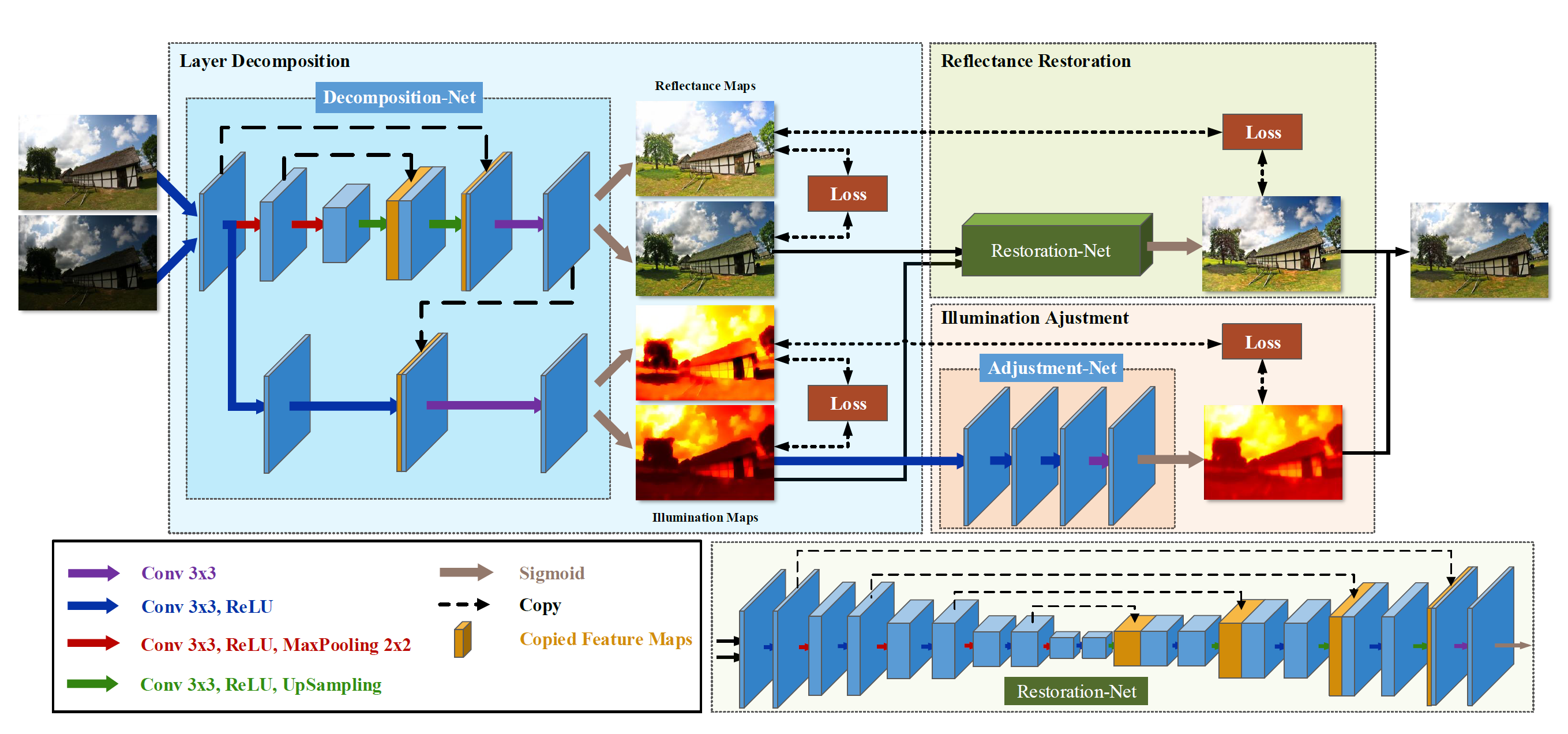}
	\end{center}
	\caption{The architecture of our KinD network. Two branches correspond to the reflectance and illumination, respectively. From the perspective of functionality, it also can be divided into three modules, including layer decomposition, reflectance restoration, and illumination adjustment.}
	\vspace{-0pt}
	\label{fig:net}
\end{figure*}

\textbf{Image Denoising Methods.} In the fields of image processing, multimedia, and computer vision,  image denoising has been a hot topic for a long time, with numerous techniques proposed over past decades. Classic ones model/regularize the problem by utilizing some specific priors of natural clean images, like non-local self-similarity, piecewise smoothness, signal (representation) sparsity, {\it etc.} The most popular schemes arguably go to BM3D \cite{BM3D} and WNNM \cite{WNNM}. {\it Due to the high complexity of optimization procedure in the testing, and the large searching space of proper parameters, these traditional methods often show the unsatisfactory performance in real situations.} Lately, deep learning based denoisers exhibit the superiority on the task. The representative works, such as SSDA using stacked sparse denoising auto-encoders \cite{SSDA1,SSDA2}, TNRD by trainable nonlinear reaction diffusion \cite{TNRD}, DnCNN with residual learning and batch normalization \cite{DnCNN}, can save computational expense thanks to only feed-forward convolution operations involved in the testing phase. However, these deep models still have the difficulty for blind image denoising. One may train multiple models for varied levels or one model with a large number of parameters, which is obviously inflexible in practice. By taking the recurrent thought into the task, this issue is mitigated \cite{Unfolding}. But, {\it none of the mentioned approaches considers that different regions of a light-enhanced image host different levels of noise. Same problem happens to color distortion.}

\subsection{Our Contributions}
This study presents a deep network for practically solving the low-light enhancement problem. The main contributions of this work can be summarized in the following aspects.
\emph{\begin{itemize}
		\item Inspired by Retinex theory, the proposed network decomposes images into two components, {\it i.e.} reflectance and illumination, which decouples the original space into two smaller ones. 
		\item The network is trained with paired images captured under different light/exposure conditions, instead of using any ground-truth reflectance and illumination information.
		\item Our designed model provides a mapping function for flexibly adjusting light levels according to different demands from users.
		\item The proposed network also contains a module, which is capable to effectively remove visual defects amplified through lightening dark regions. 
		\item Extensive experiments are conducted to demonstrate the efficacy of our design and its superiority over state-of-the-art alternatives.
\end{itemize}}

\section{Methodology}
A desired low-light image enhancer should be capable to effectively remove the degradations hidden in the darkness, and flexibly adjust light/exposure conditions. We build a deep network, denoted as KinD, to achieve the goal. As schematically illustrated in Figure \ref{fig:net}, the network is composed of two branches for handling the reflectance and illumination components, respectively. From the perspective of functionality, it also can be divided into three modules, including layer decomposition, reflectance restoration, and illumination adjustment. In the next subsections, we shall explain the details about the network.

\subsection{Consideration \& Motivation}
\label{sec:cnm}
\subsubsection{Layer Decomposition} 
As discussed in Sec. \ref{sec:PA}, the main drawback of plain methods comes from the blindness of illumination. Thus, it is key to obtain the illumination information. If having the illumination well-extracted from the input, the rest hosts the details and possible degradations, where the restoration (or degradation removal) can be executed on. In Retinex theory, an image $\bs{I}$ can be viewed as a composition of two components, {\it i.e.} reflectance $\bs{R}$ and illumination $\bs{L}$, in the fashion of $\bs{I}=\bs{R}\circ\bs{L}$, where $\circ$ designates the element-wise product. Further, decomposing images in the Retinex manner consequently decouples the space of mapping a degraded low-light image to a desired one into two smaller subspaces, expecting to be better and easier regularized/learned. Moreover, the illumination map is core to flexibly adjusting light/exposure conditions. Based on the above, {\it the Retinex-based layer decomposition is suitable and necessary for the target task.}  % From the functional perspective, the Retinex model is suitable for the target task. , which is typically of strong structure {\it e.g.} piece-wise smoothness

\subsubsection{Data Usage \& Priors} 
{\it There is no well-defined ground-truth for light conditions. Furthermore, no/few ground-truth reflectance and illumination maps for real images are available.} The layer decomposition problem is in nature under-determined, thus additional priors/regularizers matter. Suppose that the images are degradation-free, {\it different shots of a certain scene should share the same reflectance}. While the illumination maps, though could be intensively varied, {\it are of simple and mutually consistent structure}. In real situations, the degradations embodied in low-light images are often worse than those in brighter ones, which will be diverted into the reflectance component. This inspires us that the reflectance from the image in bright light can perform as the reference (ground-truth) for that from the degraded low-light one to learn restorers. One may ask that {\it why not use synthetic data?} Because it is hard to synthesize. The degradations are not in a simple form, and change with respect to different sensors. {\it Please notice that the usage of reflectance (well-defined) totally differs from using images in (relatively) bright light as the reference of low light ones.}

\subsubsection{Illumination Guided Reflectance Restoration}
{\it In the decomposed reflectance, the pollution of regions corresponding to darker illumination is heavier than that to brighter one.} Mathematically, a degraded low-light image can be naturally modeled as $\bs{I}=\bs{R}\circ\bs{L}+\bs{E}$, where $\bs{E}$ designates the pollution component. By taking simple algebra steps, we have:
\begin{equation}
\begin{aligned}
\bs{I}=\bs{R}\circ\bs{L}+\bs{E}=\tilde{\bs{R}}\circ\bs{L}=(\bs{R}+\tilde{\bs{E}})\circ\bs{L}=\bs{R}\circ\bs{L}+\tilde{\bs{E}}\circ\bs{L},
\end{aligned}
\end{equation}
where $\tilde{\bs{R}}$ stands for the polluted reflectance, and $\tilde{\bs{E}}$  is the degradation having the illumination decoupled. The relationship $\bs{E}=\tilde{\bs{E}}\circ\bs{L}$ holds. Taking the additive white Gaussian noise $\bs{E}\sim\mc{N}(0, \sigma^2)$ for an example, the distribution of $\tilde{\bs{E}}$ becomes much more complex and strongly relates to $\bs{L}$, {\it i.e.} $\frac{\sigma^2}{\bs{L}_{i}}$ for each position $i$. This is to say, the reflectance restoration cannot be uniformly processed over an entire image, and the illumination map can be a good guider. One may wonder {\it what if directly removing $\bs{E}$ from the input $\bs{I}$?} For one thing, the unbalance issue still remains. By viewing from another point, the intrinsic details will be unequally confounded with the noise. For another thing, different from the reflectance, we no longer have proper references for degradation removal in this manner, since $\bs{L}$ varies. Analogous analysis serves other types of degradation, like color-distortion. %Thus, the selection of parameter $\sigma^2$ is very sensitive. , where $v$ is the value with respect to the variable $u$ On the other hand, the restoration turns fragile as in low-light regions the intensities of both intrinsic details and noise are weak, and different light conditions are not comparative.

\subsubsection{Arbitrary Illumination Manipulation}
{\it The favorite illumination strengths of different persons/applications may be pretty diverse.} Therefore, a practical system needs to provide an interface for arbitrary illumination manipulation. In the literature, three main ways for enhancing light conditions are fusion, light level appointment, and gamma correction. The fusion-based methods, due to the fixed fusion mode, lack in the functionality of light adjustment. If adopting the second option, the training dataset has to contain images with target levels, limiting its flexibility. For gamma correction, although it can achieve the goal by setting different $\gamma$ values, it may be unable to reflect the relationship between different light (exposure) levels. {\it This paper advocates to learn a flexible mapping function from real data, which accepts users to appoint arbitrary levels of light/exposure.}

\begin{figure}[t]
	\begin{center}
			\includegraphics[width=1\linewidth]{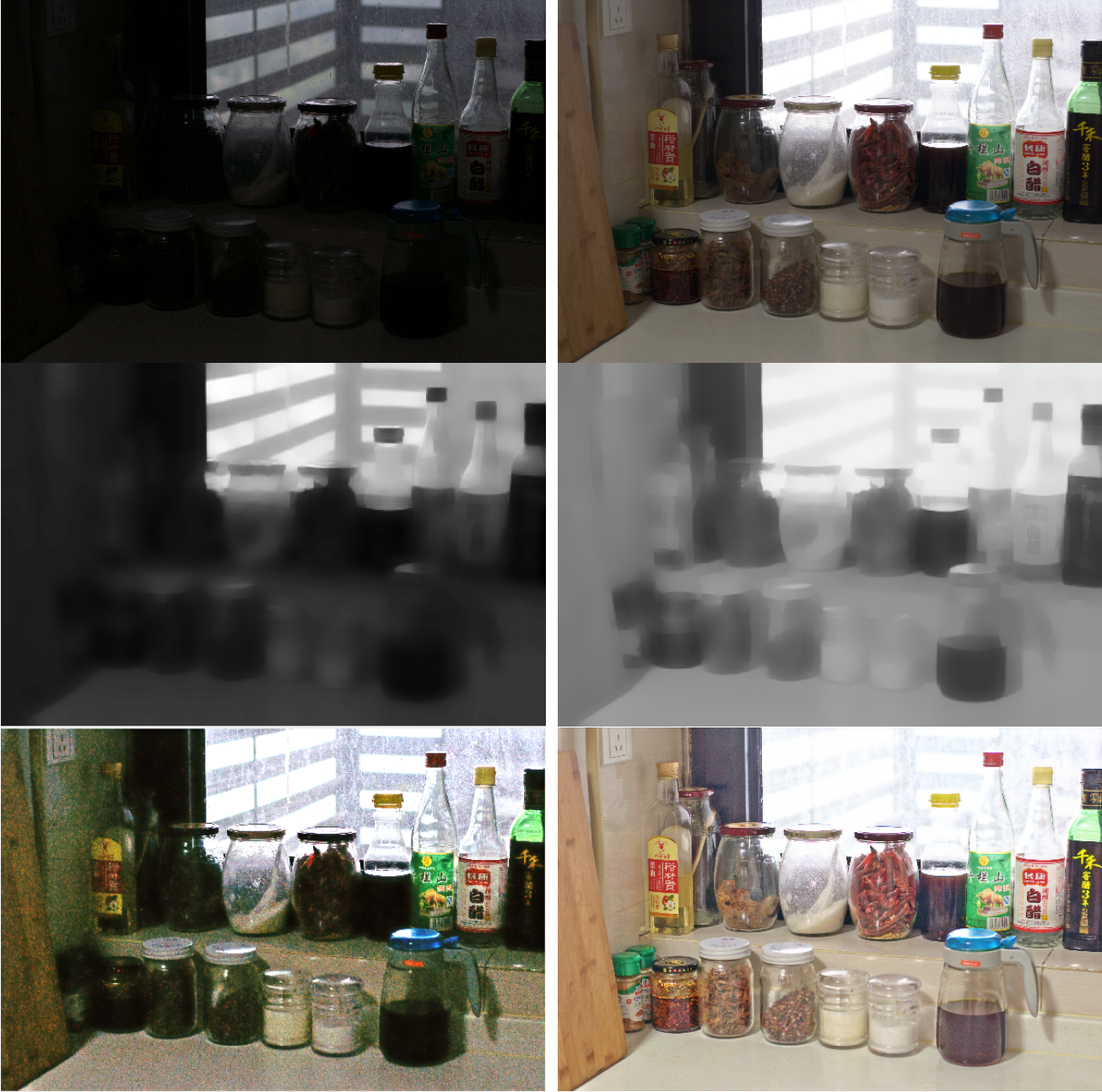}
	\end{center}
	\caption{Left column: Lower light input and its decomposed illumination and (degraded) reflectance maps. Right column: Brighter input and its corresponding maps. Three rows respectively correspond to inputs, illumination maps, and reflectance maps. These are testing images.}
	\vspace{-0pt}
	\label{fig:LD}
	
	\begin{center}
		\includegraphics[width=1\linewidth]{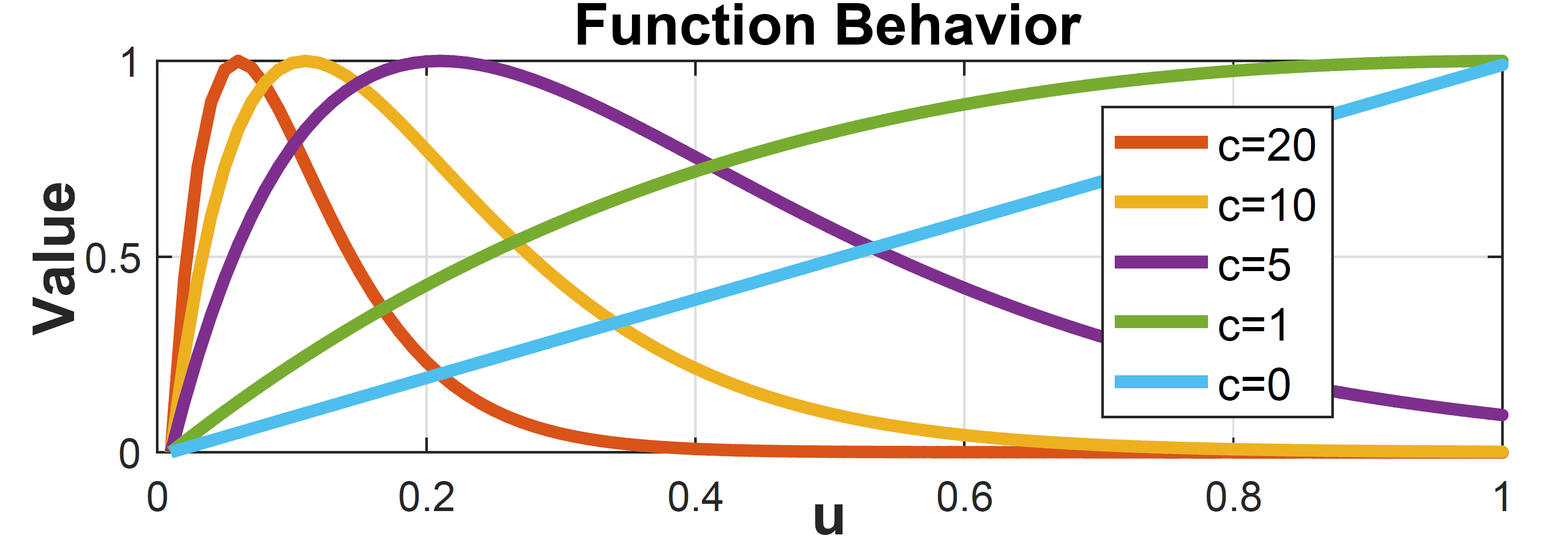}
	\end{center}
	\caption{The behavior of function $v=u\cdot\exp{(-c\cdot u)}$. The parameter $c$ controls the shape of function.}%, where $v$ is the value with respect to the variable $u$
	\vspace{-0pt}
	\label{fig:fb}
\end{figure}

\begin{table}
	\centering
	\resizebox{0.48\textwidth}{!}{
		\begin{tabular}{cccccc}
			\hline
			Inputs &Operator& Kernel &Output Channels & Stride & Output Name \\ 
			\hline
			RGB & Conv\&ReLU & $3\times3$ & 32 & 1 & Decom\_conv1 \\
			Decom\_conv1 & Max Pooling & $2\times2$ & 32 & 2 & Decom\_pool1 \\
			Decom\_pool1 & Conv\&ReLU & $3\times3$ & 64 & 1 & Decom\_conv2 \\
			Decom\_conv2 & Max Pooling & $2\times2$ & 64 & 2 & Decom\_pool2 \\
			Decom\_pool2 & Conv\&ReLU & $3\times3$ & 128 & 1 & Decom\_conv3 \\
			Decom\_conv3 & Deconv & $2\times2$ & 64 & 2 & Decom\_up1 \\
			Decom\_up1, Decom\_conv2 & Concat & - & 128 & - & Decom\_concat1 \\
			Decom\_concat1 & Conv\&ReLU & $3\times3$ & 64 & 1 & Decom\_conv4 \\
			Decom\_conv4 & Deconv & $2\times2$ & 32 & 2 & Decom\_up2 \\
			Decom\_up2, Decom\_conv1 & Concat & - & 64 & - & Decom\_concat2 \\
			Decom\_concat2 & Conv\&ReLU & $3\times3$ & 32 & 1 & Decom\_conv5 \\
			Decom\_conv5 & Conv & $3\times3$ & 3 & 1 & Decom\_conv6 \\
			Decom\_conv6 & Sigmoid & - & 3 & - & Decom\_Reflectance \\
			Decom\_conv1 & Conv\&ReLU & $3\times3$ & 32 & 1 & Decom\_i\_conv1 \\       
			Decom\_i\_conv1, Decom\_conv5 & Concat & - & 64 & - & Decom\_i\_conv2 \\
			Decom\_i\_conv2 & Conv & $3\times3$ & 1 & 1 & Decom\_i\_conv3 \\
			Decom\_i\_conv3 & Sigmoid & - & 1 & - & Decom\_Illumination \\
			\hline
		\end{tabular}
	}
	\caption{Layer decomposition network}
	\label{tab:LD}
\end{table}

\begin{table}
		\label{tab:RR}
	\centering
	\resizebox{0.48\textwidth}{!}{
		\begin{tabular}{cccccc}
			\hline
			Inputs &Operator& Kernel &Output Channels & Stride & Output Name \\
			\hline
			Decom\_i\_conv3, Decom\_conv5 & Concat & - & 33 & - & RE\_concat1 \\
			RE\_concat1 & Conv\&ReLU & $3\times3$ & 32 & 1 & RE\_conv1\_1 \\
			RE\_conv1\_1 & Conv\&ReLU & $3\times3$ & 32 & 1 & RE\_conv1\_2 \\
			RE\_conv1\_2 & Max Pooling & $2\times2$ & 32 & 2 & RE\_pool1 \\
			RE\_pool1 & Conv\&ReLU & $3\times3$ & 64 & 1 & RE\_conv2\_1 \\
			RE\_conv2\_1 & Conv\&ReLU & $3\times3$ & 64 & 1 & RE\_conv2\_2 \\
			RE\_conv2\_2 & Max Pooling & $2\times2$ & 64 & 2 & RE\_pool2 \\
			RE\_pool2 & Conv\&ReLU & $3\times3$ & 128 & 1 & RE\_conv3\_1 \\
			RE\_conv3\_1 & Conv\&ReLU & $3\times3$ & 128 & 1 & RE\_conv3\_2 \\
			RE\_conv3\_2 & Max Pooling & $2\times2$ & 128 & 2 & RE\_pool3 \\
			RE\_pool3 & Conv\&ReLU & $3\times3$ & 256 & 1 & RE\_conv4\_1 \\
			RE\_conv4\_1 & Conv\&ReLU & $3\times3$ & 256 & 1 & RE\_conv4\_2 \\
			RE\_conv4\_2 & Max Pooling & $2\times2$ & 256 & 2 & RE\_pool4 \\
			RE\_pool4 & Conv\&ReLU & $3\times3$ & 512 & 1 & RE\_conv5\_1 \\
			RE\_conv5\_1 & Conv\&ReLU & $3\times3$ & 512 & 1 & RE\_conv5\_2 \\
			RE\_conv5\_2 & Deconv & $2\times2$ & 256 & 2 & RE\_up1 \\
			RE\_up1, RE\_conv4\_2 & Concat & - & 512 & - & RE\_concat2 \\
			RE\_concat2 & Conv\&ReLU & $3\times3$ & 256 & 1 & RE\_conv6\_1 \\
			RE\_conv6\_1 & Conv\&ReLU & $3\times3$ & 256 & 1 & RE\_conv6\_2 \\
			RE\_conv6\_2 & Deconv & $2\times2$ & 128 & 2 & RE\_up2 \\
			RE\_up2, RE\_conv3\_2 & Concat & - & 256 & - & RE\_concat3 \\
			RE\_concat3 & Conv\&ReLU & $3\times3$ & 128 & 1 & RE\_conv7\_1 \\
			RE\_conv7\_1 & Conv\&ReLU & $3\times3$ & 128 & 1 & RE\_conv7\_2 \\
			RE\_conv7\_2 & Deconv & $2\times2$ & 64 & 2 & RE\_up3 \\
			RE\_up3, RE\_conv2\_2 & Concat & - & 128 & - & RE\_concat4 \\
			RE\_concat4 & Conv\&ReLU & $3\times3$ & 64 & 1 & RE\_conv8\_1 \\
			RE\_conv8\_1 & Conv\&ReLU & $3\times3$ & 64 & 1 & RE\_conv8\_2 \\
			RE\_conv8\_2 & Deconv & $2\times2$ & 32 & 2 & RE\_up4 \\
			RE\_up4, RE\_conv1\_2 & Concat & - & 64 & - & RE\_concat5 \\
			RE\_concat5 & Conv\&ReLU & $3\times3$ & 32 & 1 & RE\_conv9\_1 \\
			RE\_conv9\_1 & Conv\&ReLU & $3\times3$ & 256 & 1 & RE\_conv9\_2 \\
			RE\_conv9\_2 & Conv & $3\times3$ & 3 & 1 & RE\_conv10 \\
			RE\_conv10 & Sigmoid & - & 3 & - & RE\_refletance \\
			\hline
		\end{tabular}
	}
	\caption{Reflectance restoration network}
	\label{tab:RR}
\end{table}

\subsection{KinD Network}
Inspired by the consideration and motivation, we build a deep neural network, denoted as KinD, for kindling the darkness. Below, we describe the three subnets in details from the functional perspective. 
\subsubsection{Layer Decomposition Net}

Recovering two components from one image is a highly ill-posed problem. Having no ground-truth information guided, a loss with well-designed constraints is important. Fortunately, we have paired images with different light/exposure configurations [$\bs{I}_l$, $\bs{I}_h$]. Recall that the reflectance of a certain scene should be shared across different images, we regularize the decomposed reflectance pair [$\bs{R}_l$, $\bs{R}_h$] to be close (ideally the same if degradation-free). Furthermore, the illumination maps [$\bs{L}_l$, $\bs{L}_h$] should be piece-wise smooth and mutually consistent. The following terms are adopted. We simply use $\mc{L}_{rs}^{LD}\defeq\|\bs{R}_l-\bs{R}_h\|_2^2$ to regularize the {\it reflectance similarity}, where $\|\cdot\|_2$ means the $\ell^2$ norm (MSE). The {\it illumination smoothness} is constrained by $\mc{L}_{is}^{LD}\defeq\|\frac{\nabla\bs{L}_l}{max(|\nabla\bs{I}_l|,\epsilon)}\|_1+\|\frac{\nabla\bs{L}_h}{max(|\nabla\bs{I}_h|,\epsilon)}\|_1$, where $\nabla$ stands for the first order derivative operator containing $\nabla_x$ (horizontal) and $\nabla_y$ (vertical) directions, and $\|\cdot\|_1$ means the $\ell^1$ norm. In addition, $\epsilon$ is a small positive constant (0.01 in this work) for avoiding zero denominator, and $|\cdot|$ means the absolute value operator. This smoothness term measures the relative structure of the illumination with respect to the input. For a location on an edge in $\bs{I}$, the penalty on $\bs{L}$ is small; while for a location in a flat region in $\bs{I}$, the penalty turns to be large. As for the {\it mutual consistency}, we employ $\mc{L}_{mc}^{LD}\defeq\|\bs{M}\circ\exp(-c\cdot\bs{M})\|_1$ with $\bs{M}\defeq|\nabla\bs{L}_l|+|\nabla\bs{L}_h|$. Figure \ref{fig:fb} depicts the function behavior of $u\cdot\circ\exp(-c\cdot u)$, where $c$ is the parameter controlling the shape of function. As can be seen from Figure \ref{fig:fb}, the penalty first goes up but then drops towards $0$ as $u$ increases. This characteristic well fits the mutual consistency, {\it i.e.} strong mutual edges should be preserved while weak ones depressed. We notice that setting $c=0$ leads to a simple $\ell^1$ loss on $\bs{M}$. Besides, the decomposed two layers should reproduce the input, which is constrained by the {\it reconstruction error}, say $\mc{L}_{rec}^{LD}\defeq\|\bs{I}_l-\bs{R}_l\circ\bs{L}_l\|_1+\|\bs{I}_h-\bs{R}_h\circ\bs{L}_h\|_1$. As a result, the loss function of layer decomposition net is as follows:
\begin{equation}
\mc{L}^{LD}\defeq\mc{L}_{rec}^{LD}+0.01\mc{L}_{rs}^{LD}+0.08\mc{L}_{is}^{LD}+0.1\mc{L}_{mc}^{LD}.
\end{equation}  
The layer decomposition network contains two branches corresponding to the reflectance and illumination, respectively. The reflectance branch adopts a typical 5-layer U-Net \cite{UNet}, followed by a convolutional (conv) layer and a Sigmoid layer. While the illumination branch is composed of two conv+ReLU layers and a conv layer on concatenated feature maps from the reflectance branch (for possibly excluding textures from the illumination), finally followed by a Sigmoid layer. The detailed layer decomposition network configuration is provided in Table \ref{tab:LD}. % Due to page limit,  supplementary material. Illustratively, Figure \ref{fig:LD} offers an example of the layer decomposition network, from which we can see that the separation is very promising (structure preserved illumination and rich reflectance). Note that the reflectance of the lower light one contains noise and color distortion, which will be removed by the reflectance restoration net.   

\begin{figure}[t]
	
	\begin{center}
			\includegraphics[width=1\linewidth]{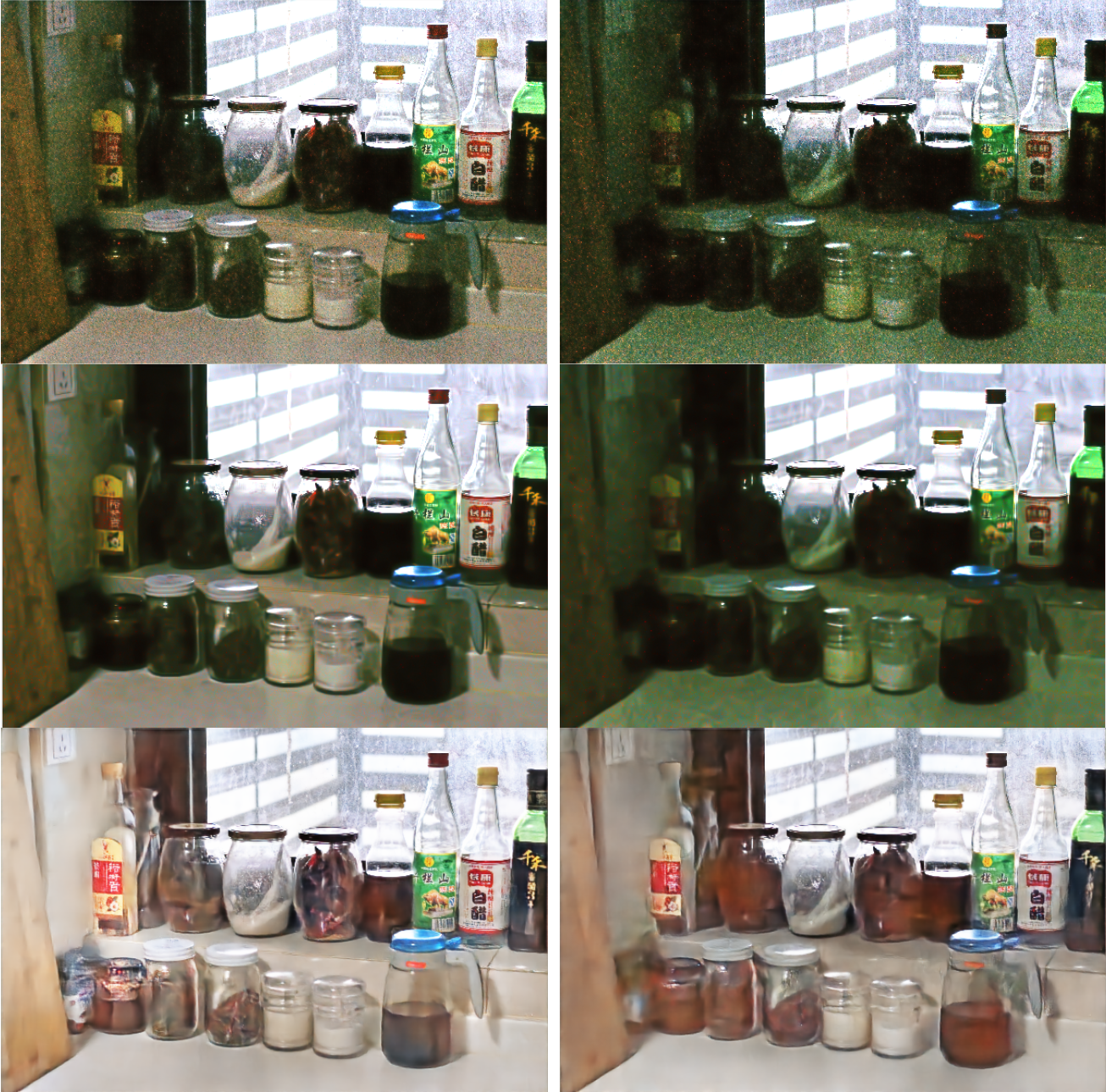}
	\end{center}
	\caption{The polluted reflectance maps (top), and their results by BM3D (middle) and our reflectance restoration net (bottom). The right column corresponds to a heavier degradation (a lower light) level than the left. These are testing images.}
	\vspace{-0pt}
	\label{fig:RR}
\end{figure}

\begin{figure*}[t]
	
	\begin{center}
			\includegraphics[width=1\linewidth]{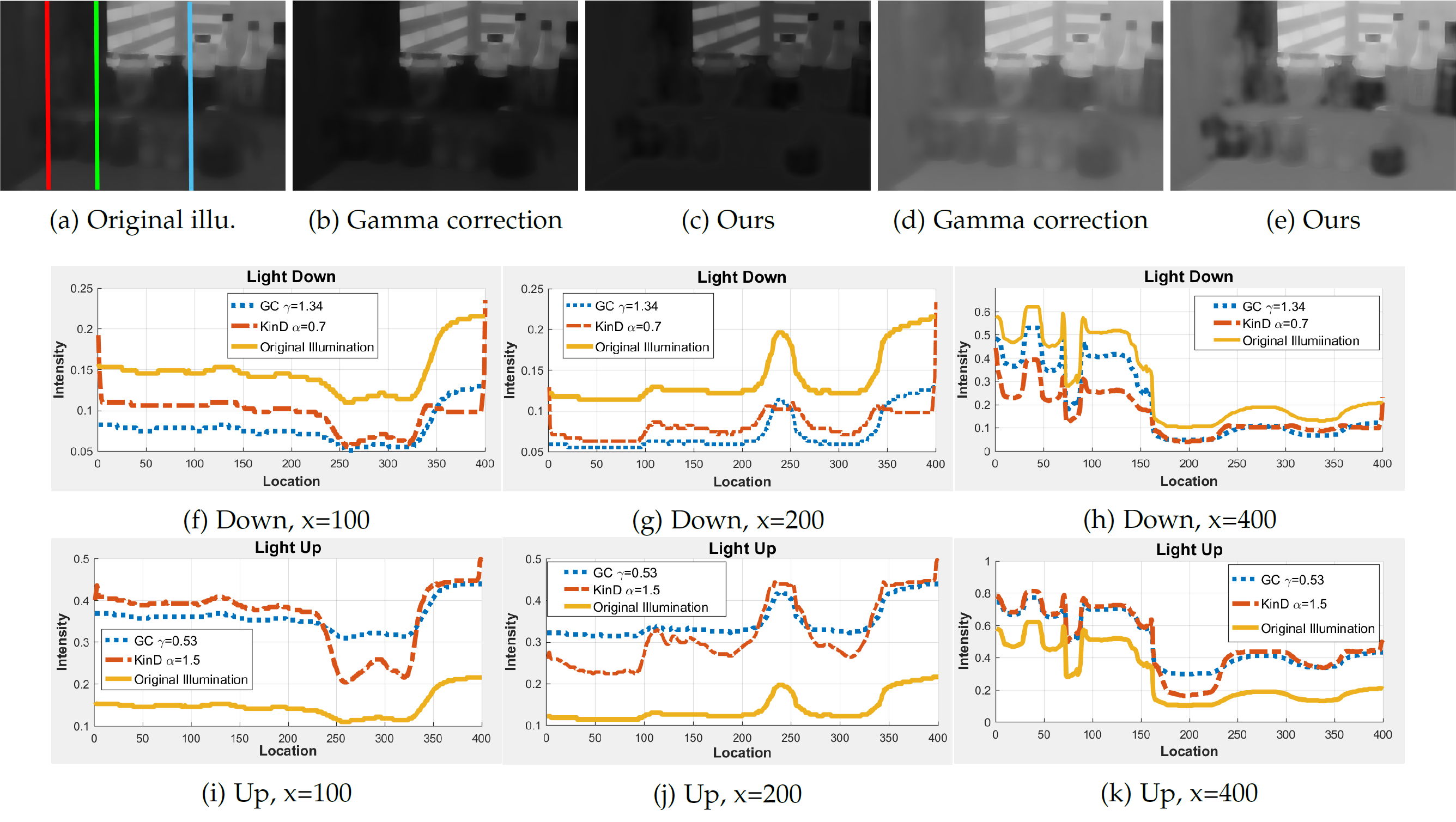}
	\end{center}
	\caption{Comparison between Gamma correction and our illumination adjustment manner. (a) shows the original/source illumination map. Two cases, including 1) turning the light down with $\gamma=1.34$ (b) and $\alpha=0.7$ (c), and 2) turning the light up with $\gamma=0.53$ (d) and $\alpha=1.5$ (e), are provided. (f)-(k) give the 1D curves at $x=100, 200, 400$ corresponding to the red, green, and blue lines in (a), respectively.}
	\vspace{-0pt}
	\label{fig:IA}
\end{figure*}

\subsubsection{Reflectance Restoration Net}
The reflectance maps from low-light images, as shown in Figures \ref{fig:LD} and \ref{fig:RR}, are more interfered by degradations than those from bright-light ones. Employing the clearer reflectance to act as the reference (informal ground-truth) for the messy one is our principle. For seeking a restoration function, the objective turns to be simple as follows:
\begin{equation}
\mc{L}^{RR}\defeq \|\hat{\bs{R}}-\bs{R}_h\|_2^2- \text{SSIM}(\hat{\bs{R}},\bs{R}_h)+\|\nabla\hat{\bs{R}}-\nabla\bs{R}_h\|_2^2,
\end{equation}
where $\text{SSIM}(\cdot,\cdot)$ is the structural similarity measurement, and $\hat{\bs{R}}$ corresponds to the restored reflectance. The third term concentrates on the closeness in terms of textures. This subnet is similar to the reflectance branch in the layer decomposition subnet, but deeper. The schematic configuration is given in Figure \ref{fig:net} and detailed in Appendix. %supplementary material. 
We recall that {\it the degradation distributes in the reflectance complexly, which strongly depends on the illumination distribution}. Thus, we bring the illumination information into the restoration net together with the degraded reflectance. The effectiveness of this operation can be observed in Figure \ref{fig:RR}. In the two reflectance maps with different degradation (light) levels, the results by BM3D can fairly remove noise (without regarding the color distortion in nature). The blur effect exists almost everywhere. In our results, the textures (the dust/water-based stains for example) of the window region, which is originally bright and barely polluted, keeps clear and sharp, while the degradations in the dark region get largely removed with details ({\it e.g.} the characters on the bottles) very well maintained. Besides, the color distortion is also cured by our method.  The detailed reflectance restoration network configuration is provided in Table \ref{tab:RR}.

\begin{figure*}[p]
	\begin{center}
			\includegraphics[width=1\linewidth]{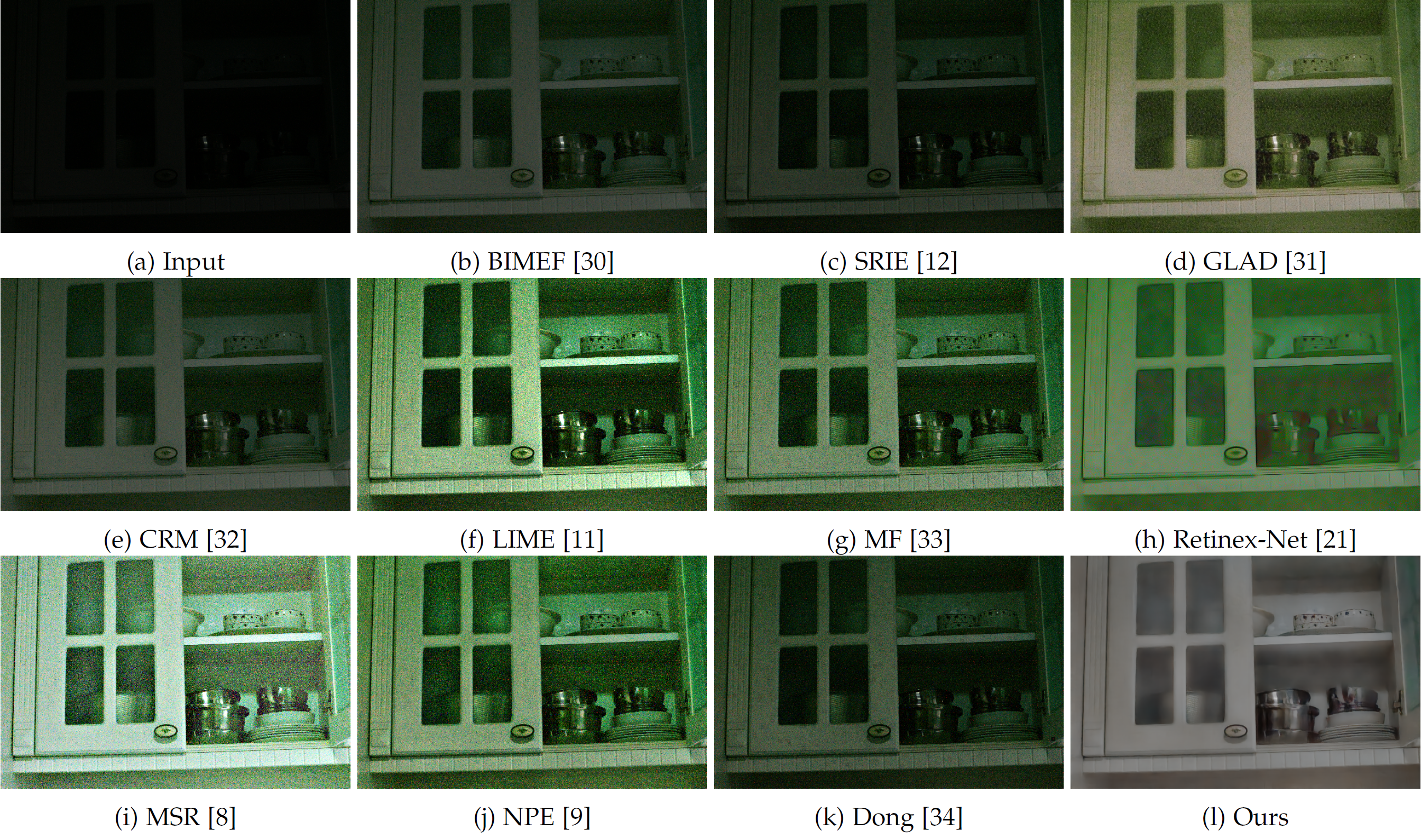}
	\end{center}
	\caption{Visual comparison with state-of-the-art low-light image enhancement methods.}
	\vspace{-0pt}
	\label{fig:comp}
	
	\begin{center}
		\includegraphics[width=1\linewidth]{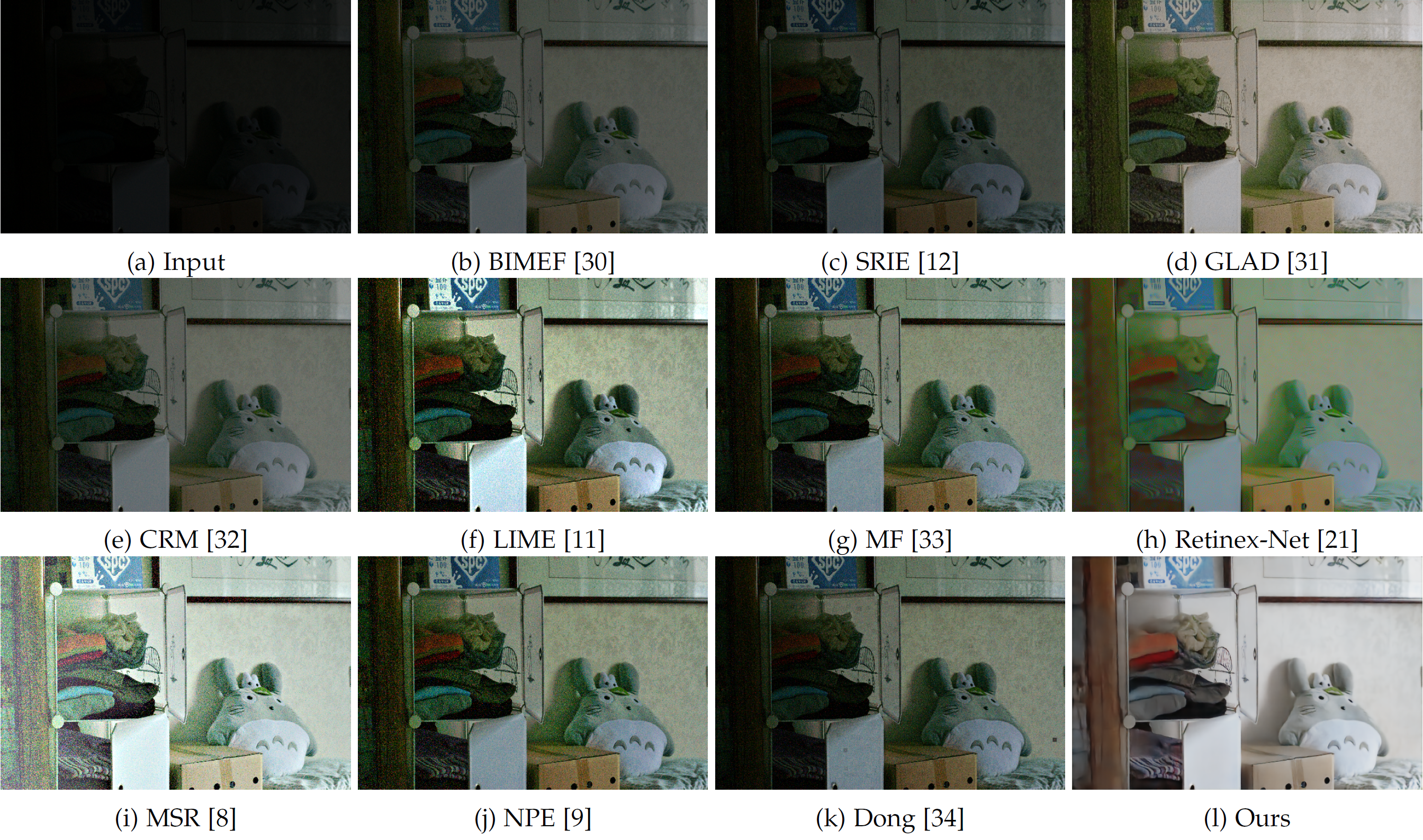}
	\end{center}
	\caption{Visual comparison with state-of-the-art low-light image enhancement methods.}
	\vspace{-0pt}
	\label{fig:comp2}
\end{figure*}
\begin{figure*}[p]
	\begin{center}
		\includegraphics[width=1\linewidth]{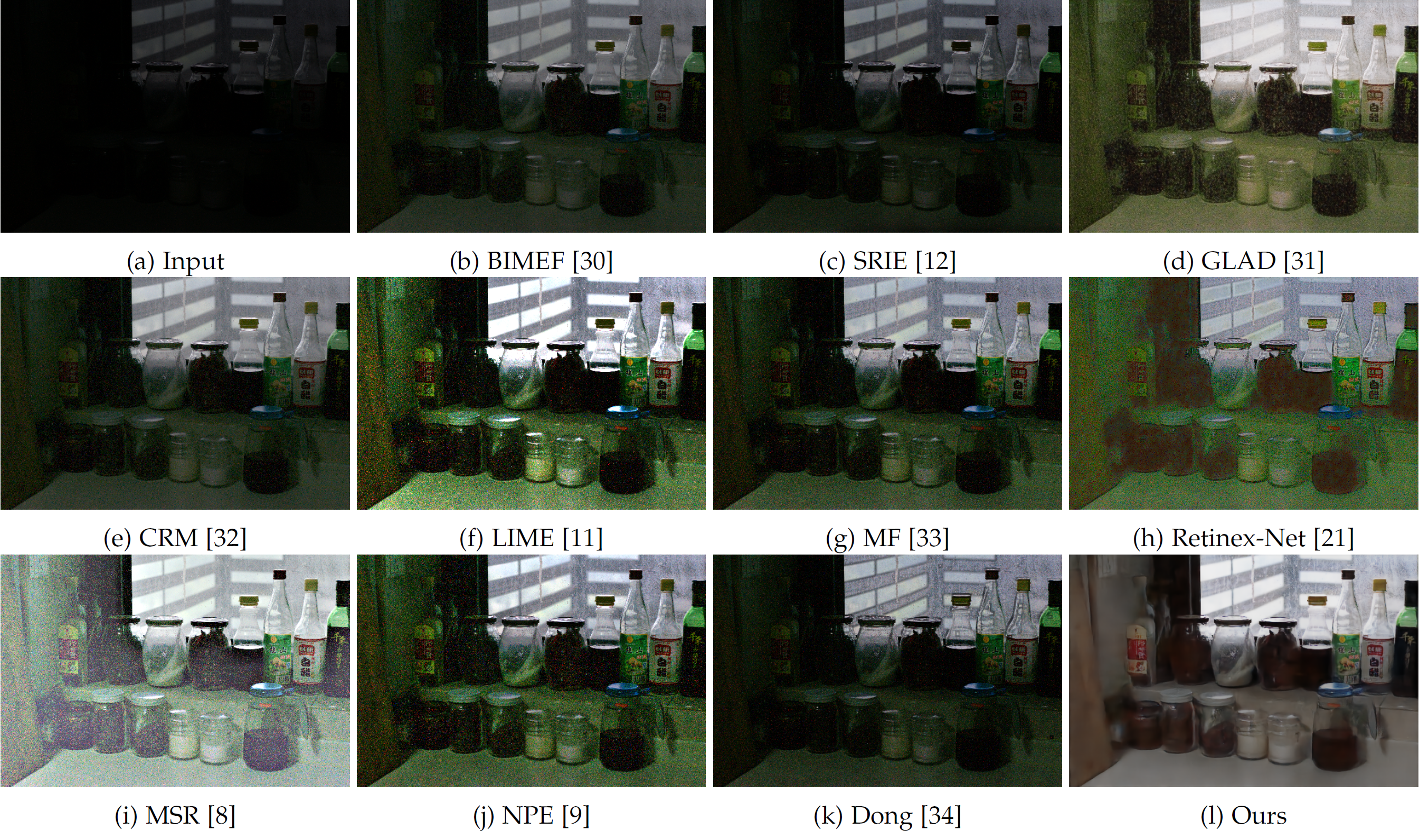}
	\end{center}
	\caption{Visual Comparison with state-of-the-art low-light image enhancement methods.}
	\vspace{-0pt}
	\label{fig:comp2}
	
	\begin{center}
			\includegraphics[width=1\linewidth]{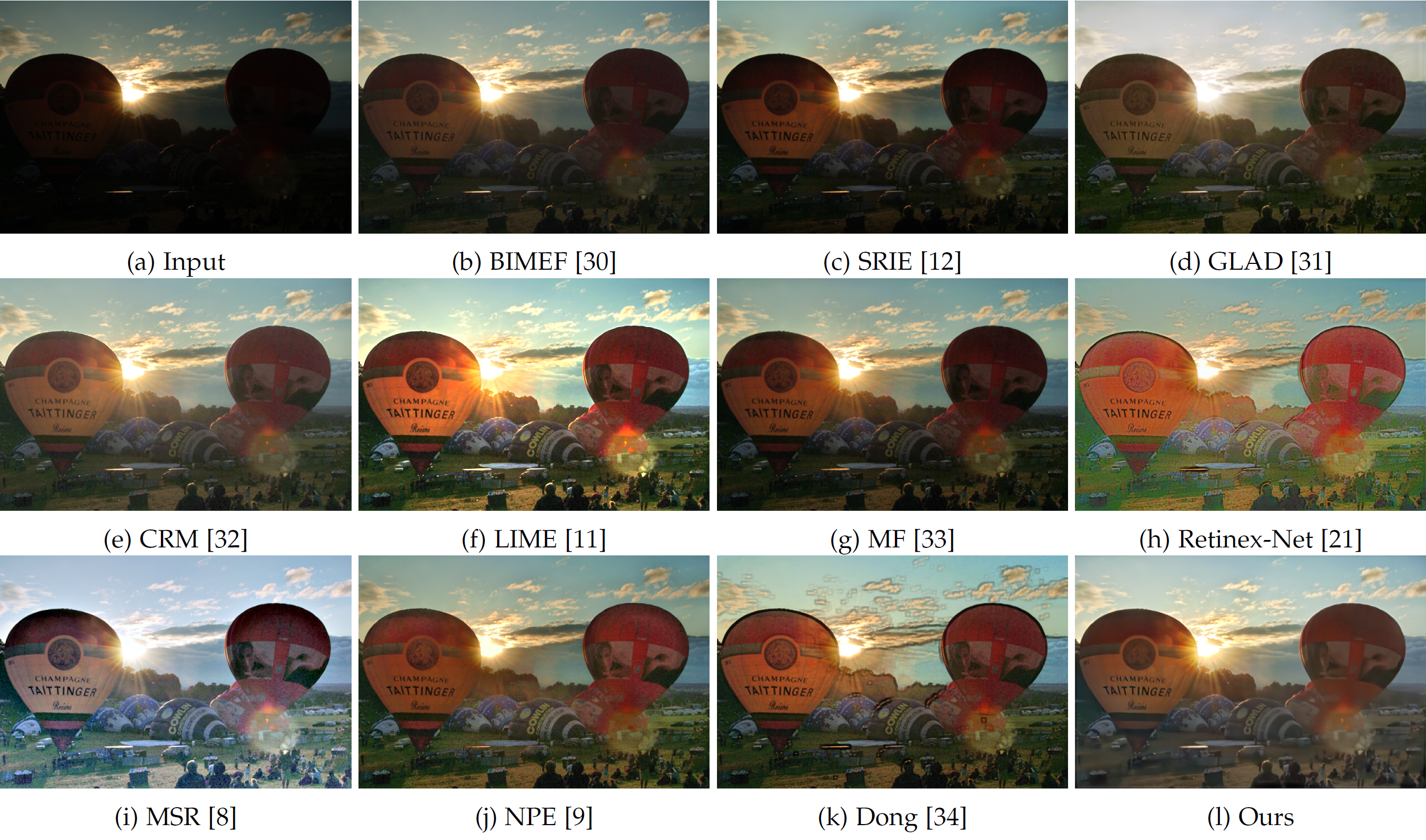}
	\end{center}
	\caption{Visual Comparison with state-of-the-art low-light image enhancement methods.}
	\vspace{-0pt}
	\label{fig:comp2}
\end{figure*}

\begin{figure*}[t]
	\begin{center}
		\includegraphics[width=1\linewidth]{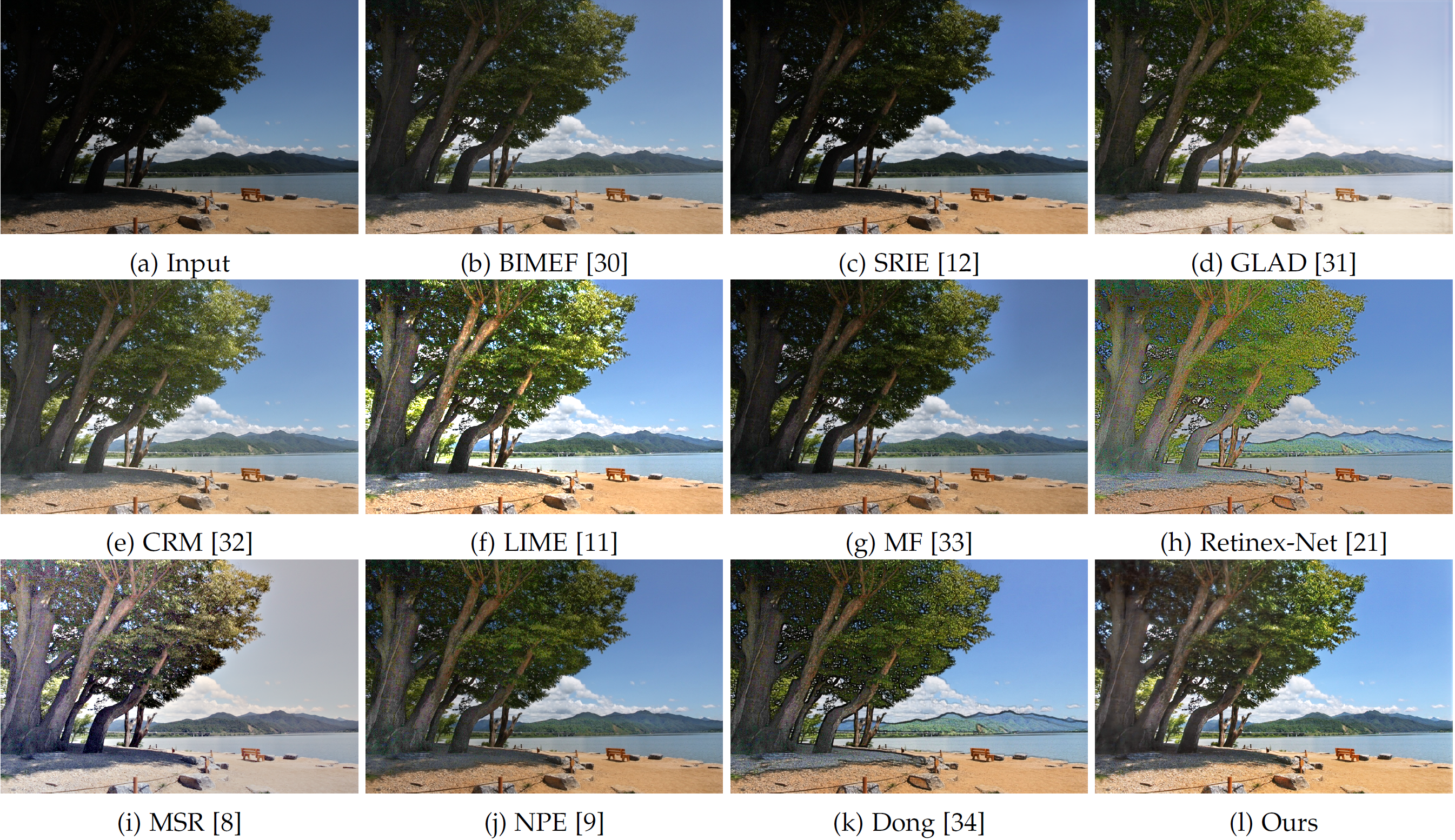}
	\end{center}
	\caption{Visual Comparison with state-of-the-art low-light image enhancement methods.}
	\vspace{-0pt}
	\label{fig:comp2}
\end{figure*}

\begin{figure*}[p]
	\begin{center}
	\includegraphics[width=1\linewidth]{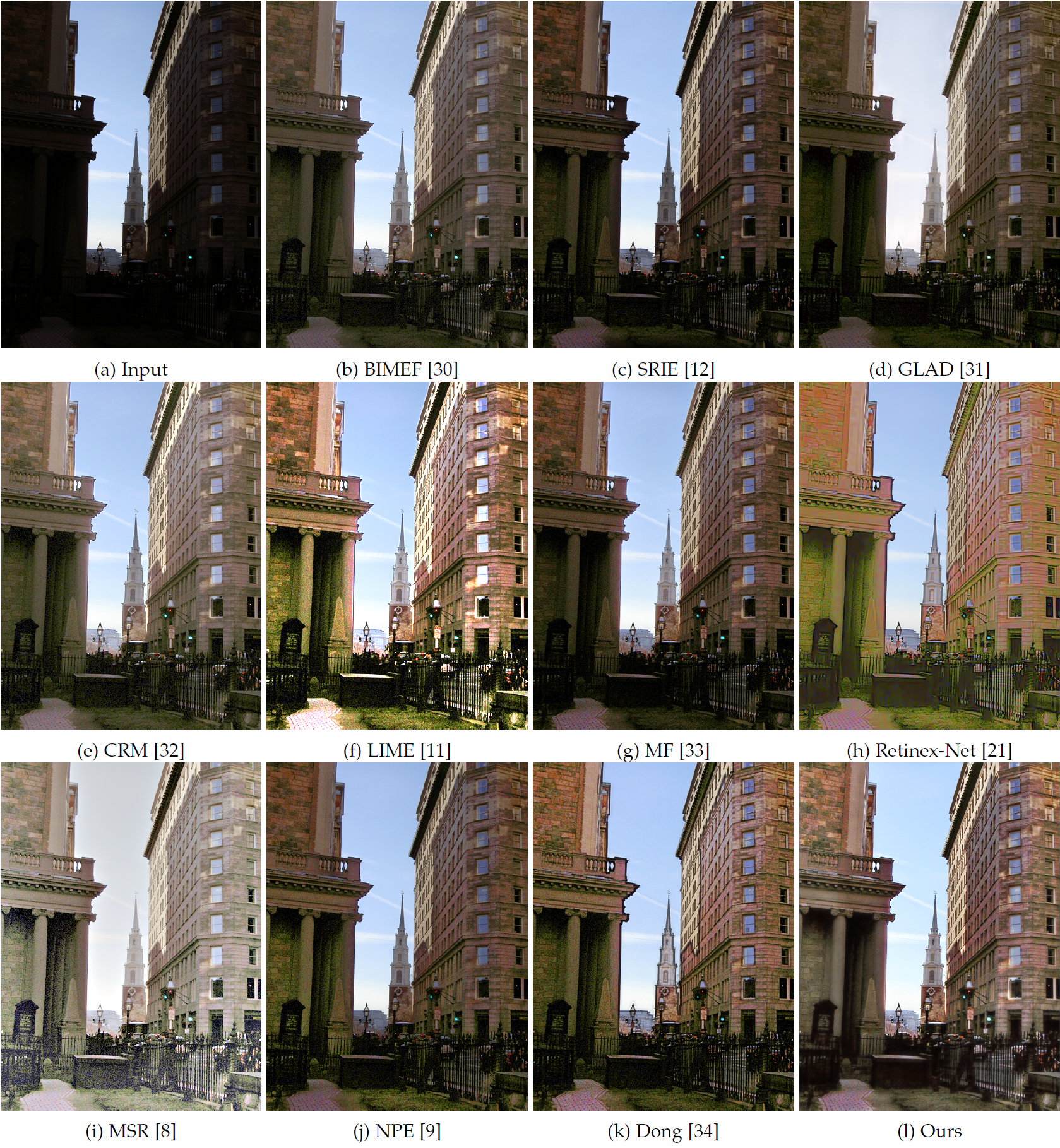}
	\end{center}
	\caption{Visual Comparison with state-of-the-art low-light image enhancement methods.}
	\vspace{-0pt}
	\label{fig:comp2}
\end{figure*}

\begin{figure*}[p]
	\begin{center}
		\includegraphics[width=1\linewidth]{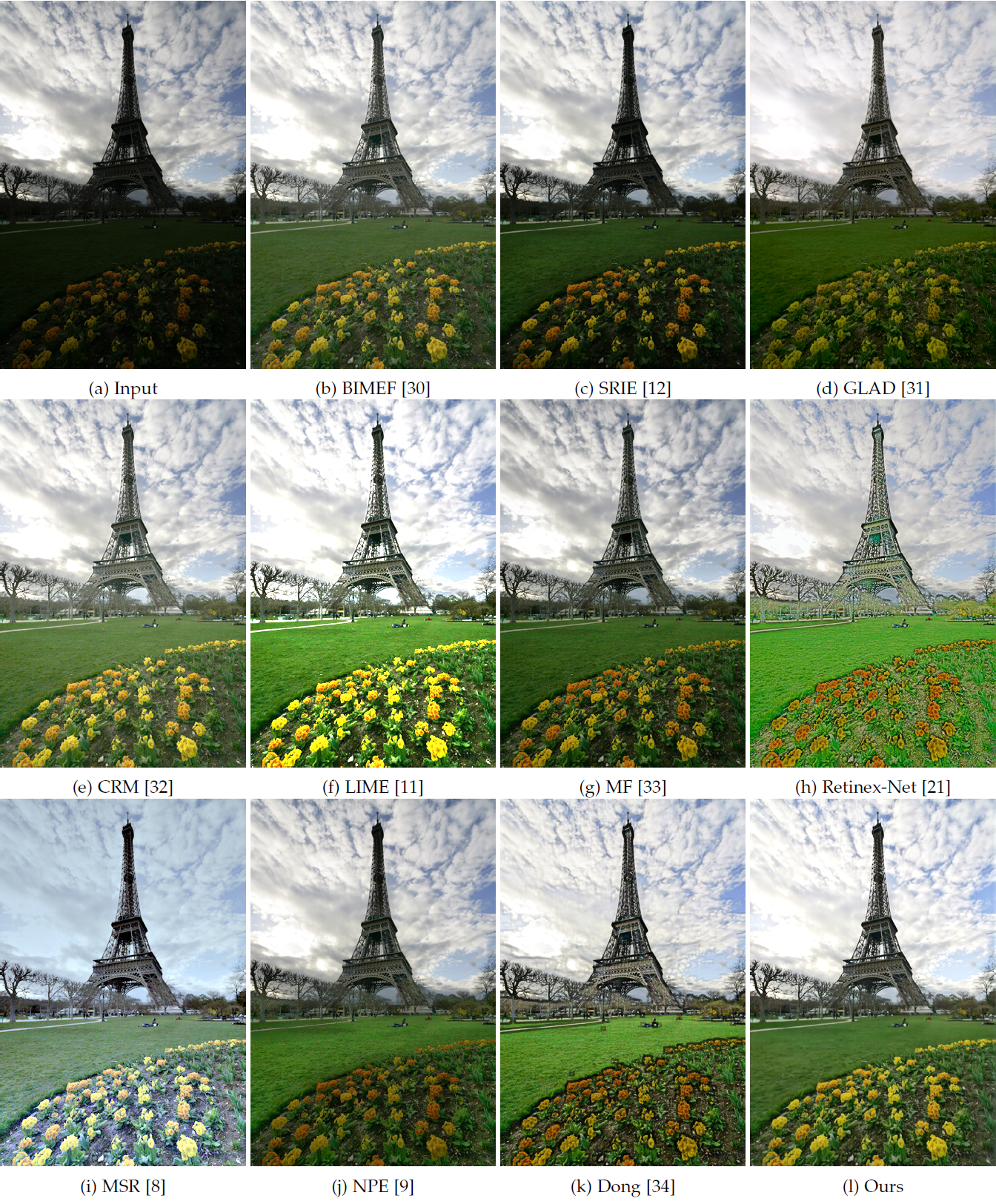}
	\end{center}
	\caption{Visual Comparison with state-of-the-art low-light image enhancement methods.}
	\vspace{-0pt}
	\label{fig:comp101}
\end{figure*}

\begin{table}
	\centering
	\resizebox{0.48\textwidth}{!}{
		\begin{tabular}{cccccc}
			\hline
			Inputs &Operator& Kernel &Output Channels & Stride & Output Name \\ 
			\hline
			Decom\_illumination, Ratio & Concat & - & 2 & - & Adjust\_concat1 \\
			Adjust\_concat1 & Conv\&ReLU & $3\times3$ & 32 & 1 & Adjust\_conv1 \\
			Adjust\_conv1 & Conv\&ReLU & $3\times3$ & 32 & 1 & Adjust\_conv2 \\
			Adjust\_conv2 & Conv\&ReLU & $3\times3$ & 32 & 1 & Adjust\_conv3 \\
			Adjust\_conv3 & Conv & $3\times3$ & 1 & 1 & Adjust\_conv4 \\
			Adjust\_conv4 & Sigmoid & - & 1 & - & Adjust\_illumination \\
			\hline
		\end{tabular}
	}
	\caption{Illumination adjustment network}
	\label{tab:IA}
\end{table}
\subsubsection{Illumination Adjustment Net}
There does not exist a ground-truth light level for images. Therefore, for fulfilling diverse requirements, we need a mechanism to flexibly convert one light condition to another. We have paired illumination maps. Even though without knowing the exact relationship between the paired illuminations, we can roughly calculate their ratio of strength, {\it i.e.} $\alpha$ by $\text{mean}(\bs{L}_t/\bs{L}_s)$ where the division is element-wise. This ratio can be used as an indicator to train an adjustment function from a source light $\bs{L}_s$ to a target one $\bs{L}_t$. If adjusting a lower level of light to a higher one, $\alpha>1$, otherwise $\alpha\leq 1$. In the testing phase, $\alpha$ can be specified by users. The network is lightweight, containing 3 conv layers (two conv+ReLu, and one conv) and 1 Sigmoid layer. We notice that {\it the indicator $\alpha$ is expanded to a feature map, acting as a part of input for the net.} The following is the loss for illumination adjustment net:  
\begin{equation}
\mc{L}^{IA}\defeq \|\hat{\bs{L}}-\bs{L}_t\|_2^2+\||\nabla\hat{\bs{L}}|-|\nabla\bs{L}_t|\|_2^2,
\end{equation}
where $\bs{L}_t$ can be $\bs{L}_h$ or $\bs{L}_l$, and $\hat{\bs{L}}$ is the adjusted illumination map from the source light ($\bs{L}_h$ or $\bs{L}_l$) towards the target one. Figure \ref{fig:IA} shows the difference between our learned adjustment function and gamma correction. For comparison fairness, we tune the parameter $\gamma$ for gamma correction to reach a similar overall light strength with ours via $\gamma =\frac{\|\log(\hat{\bs{L}})\|_1}{\|\log(\bs{L}_s)\|_1}$. We consider two adjustments without loss of generality, including one light down and one light up. Figure \ref{fig:IA} (a) depicts the source illumination, (b) and (d) are the adjusted results by gamma correction, while (c) and (e) are ours. To more clearly show the difference, we plot the 1D intensity curves at $x=100, 200, 400$. As for the light-down case, our learned manner decreases more than gamma correction in intensity on relatively bright regions, while less or about the same on dark regions. Regarding the light-up case, the opposite trend appears. In other words, our method increases less the light on relatively dark regions, while more or about the same on bright regions. The learned manner is more corroborative with actual situations. Furthermore, the $\alpha$ fashion is more convenient than the $\gamma$ way for users to manipulate. For instance, setting $\alpha$ to 2 means turns the light 2x up. The detailed illumination adjustment network configuration is provided in Table \ref{tab:IA}.

\section{Experimental Validation}

%\begin{table}[t]
%	\resizebox{0.48\textwidth}{!}{
%	\begin{tabular}{c|ccc|ccc}
%		\hline
%		Metrics&    &  LOE  &     &   &  NIQE  &   \\
%		\hline
%		&LIME-data & NPE-data & MEF-data & LIME-data &NPE-data & MEF-data  \\ 
%		\hline
%		BIMEF \cite{BIMEF}   & \textbf{478.57}  & \textbf{320.27}  & \textbf{336.70}  & 3.8169  & 4.1963  & 3.4237  \\
%		CRM \cite{CRM}    & 498.701 & 515.83  & 344.49  & 3.8546  & 3.9220  & \textbf{3.2708} \\
%		Dong \cite{Dong}   & 1244.0  & 1023.1  & 1079.6  & 4.0516  & 4.1263  & 4.1094 \\
%		LIME \cite{LIME}   & 1804.8  & 1127.0  & 1073.2  & 4.1549  & 4.2629  & 3.7159 \\
%		MF \cite{MF}     & 570.50  & 929.93  & 849.74  & 4.0689  & 4.1096  & 3.4773  \\ 
%		RRM \cite{RRM}    & 1339.0  & 1186.8  & 792.39  & 4.6426  & 4.8452  & 4.1535  \\ 
%		SRIE \cite{SRIE}   & 823.61  & 540.56  & 778.42  & 3.7863  & 3.9795  & 3.4577  \\ 
%		DRD \cite{DRD}    & 1882.5  & 1235.0  & 1815.6  & 4.5977  & 4.5674  & 4.4755  \\ 
%		MSR \cite{MSR}    & 2171.9  & 1892.0  & 1675.3  & 3.7642  & 4.3663  & 3.6096  \\ 
%		NPE \cite{NPE}    & 1471.3  & 658.99  & 1205.9  & 3.9048  & 3.9520  & 3.5378  \\ 
%		GLAD \cite{GLAD}   & 534.16  & 495.14  & 460.12  & 4.1280  & 3.9699  & 3.3435  \\ 
%		KinD    & 1016.5  & 875.22  & 819.45  & \textbf{3.7236}  & \textbf{3.8826}  & \underline{\textit{3.3429}}  \\ 
%		
%		\hline
%	\end{tabular}
%}
%	\caption{Quantitative comparison on LIME, NPE, and MEF datasets in terms of LOE and NIQE.}
%\end{table}

\subsection{Implementation Details}

We use the LOL dataset as the training dataset, which includes 500 low/normal-light image pairs. In the training, we merely employ 450 image pairs, and no synthetic images are used. For the layer decomposition net, batch size is set to be 10 and patch-size to be 48x48. While for the reflectance restoration net and illumination adjustment net, batch size is set to be 4 and patch-size to be 384x384. We use the stochastic gradient descent (SGD) technique for optimization. The entire network is trained on a Nvidia GTX 2080Ti GPU and Intel Core i7-8700 3.20GHz CPU using the Tensorflow framework.

\subsection{Performance Evaluation}
We evaluate our method on widely-adopted datasets, including LOL \cite{DRD}, LIME \cite{LIME}, NPE \cite{NPE}, and MEF \cite{MEF}. Four metrics are adopted for quantitative comparison, which are PSNR, SSIM, LOE \cite{NPE}, and NIQE \cite{NIQE}. A higher value in terms of PSNR and SSIM indicates better quality, while, in LOE and NIQE, the lower the better. The state-of-the-art methods of BIMEF \cite{BIMEF}, SRIE \cite{SRIE}, CRM \cite{CRM}, Dong \cite{Dong}, LIME \cite{LIME}, MF \cite{MF}, RRM \cite{RRM}, Retinex-Net \cite{DRD}, GLAD \cite{GLAD}, MSR \cite{MSR} and NPE \cite{NPE} are involved as the competitors. 

\begin{table*}[t]
	\centering
	%	\resizebox{0.48\textwidth}{!}{
	\begin{tabular}{c|cccccc}
		\hline
		Metrics &BIMEF \cite{BIMEF} & CRM \cite{CRM} & Dong \cite{Dong} & LIME \cite{LIME} & MF \cite{MF} & RRM \cite{RRM}  \\ 
		\hline
		PSNR   & 13.8753 & 17.2033 & 16.7165 & 16.7586 & 18.7916 & 13.8765  \\
		SSIM   & 0.5771  & 0.6442  & 0.5824  & 0.5644  & 0.6422  & 0.6577 \\
		LOE    & 1456.1  & 1757.7  & \textbf{1283.2}  & 1909.5  & 2051.7  & 2025.5 \\
		LOE$_{ref}$    & 985.9  & \textbf{926.1}  & 1391.5  & 1342.4  & 1042.1  & 958.7 \\
		NIQE   & 7.5150  & 7.6865  & 8.3157  & 8.3777  & 8.8770  & 5.8101 \\
		\hline
		Metrics & SRIE \cite{SRIE} & Retinex-Net \cite{DRD} & MSR \cite{MSR} & NPE \cite{NPE} & GLAD \cite{GLAD} & KinD \\
		\hline
		PSNR   & 11.8552 & 16.7740 & 13.1728 & 16.9697 & 19.7182 & \textbf{20.8665} \\
		SSIM   & 0.4979  & 0.5594  & 0.4787  & 0.5894  & 0.7035  & \textbf{0.8022}  \\
		LOE    & 1745.4  & 2449.3  & 2589.4  & 2076.3  & 1795.5  & 2012.2 \\
		LOE$_{ref}$    & 1199.8  & 2201.7  & 2084.8  & 1643.1  & 1017.1  & \underline{977.3} \\
		NIQE   & 7.2869  & 8.8785  & 8.1136  & 8.4390  & 6.4755  & \textbf{5.1461} \\      
		
		\hline
	\end{tabular}
	%	}
	\vspace{10pt}
	\caption{Quantitative comparison on LOL dataset in terms of PSNR, SSIM, LOE, LOE$_{ref}$, and NIQE. The best results are highlighted in bold.}
	%\vspace{-10pt}
	\label{tab:LOL}
\end{table*}

Table \ref{tab:LOL} reports the numerical results among the competitors on LOL dataset. For each testing low-light image, there is a ``normal"-light correspondence. Thus, the correspondence can be taken as the reference to measure PSNR and SSIM. From the numbers, we see that our KinD significantly outperforms all the other methods. In terms of the non-reference metric NIQE, our KinD also takes the first place by a large margin. But, in LOE, our method seems falling behind many methods. As the authors of \cite{LIME} stated, using the low-light input itself to compute LOE is problematic. One should choose a reliable reference. Similar to computing PSNR and SSIM, we again employ the correspondence image as the reference (denoted as LOE$_{ref}$). In this way, our KinD comes up to the $3$rd place, slightly inferior to CRM (977.3 vs. 926.1).  
Regarding the LIME, NPE, and MEF datasets, no reference images are available. Thus, we only adopt the NIQE to evaluate the performance difference among the involved methods. In this comparison, as given in Tab. \ref{tab:other}, our KinD shows its clear advantage against the others. Specifically, KinD outperforms all the competitors on the LIME and NPE datasets. For the MEF data, it is only behind CRM by a small difference (3.34 vs. 3.27). 

\begin{table}[t]
	\centering
	%\resizebox{0.48\textwidth}{!}{
	\begin{tabular}{c|ccc}
		\hline
		Metric   &   &  NIQE  &   \\
		\hline
		Datasets& LIME-data &NPE-data & MEF-data  \\ 
		\hline
		BIMEF \cite{BIMEF}    & 3.8169  & 4.1963  & 3.4237  \\
		CRM \cite{CRM}     & 3.8546  & 3.9220  & \textbf{3.2708} \\
		Dong \cite{Dong}     & 4.0516  & 4.1263  & 4.1094 \\
		LIME \cite{LIME}    & 4.1549  & 4.2629  & 3.7159 \\
		MF \cite{MF}       & 4.0689  & 4.1096  & 3.4773  \\ 
		RRM \cite{RRM}     & 4.6426  & 4.8452  & 4.1535  \\ 
		SRIE \cite{SRIE}    & 3.7863  & 3.9795  & 3.4577  \\ 
		Retinex \cite{DRD}    & 4.5977  & 4.5674  & 4.4755  \\ 
		MSR \cite{MSR}     & 3.7642  & 4.3663  & 3.6096  \\ 
		NPE \cite{NPE}      & 3.9048  & 3.9520  & 3.5378  \\ 
		GLAD \cite{GLAD}   & 4.1280  & 3.9699  & 3.3435  \\ 
		KinD     & \textbf{3.7236}  & \textbf{3.8826}  & \underline{\textit{3.3429}}  \\ 
		
		\hline
	\end{tabular}
	%	}
	\caption{Quantitative comparison on LIME, NPE, and MEF datasets in terms of NIQE. The best results are highlighted in bold.}
	\label{tab:other}
	\vspace{-10pt}
\end{table} 

In addition, Figures \ref{fig:comp}-\ref{fig:comp101} give a number of visual comparisons on the images with different light conditions. From the results, we can see that, although most of methods can somehow brighten the inputs, severe visual defects caused by unsatisfactory adjustment of light and/or obstinate noise and color distortion remain. Our KinD works well in these cases with the light properly adjusted and degradations clearly removed. %Due to limited space, more visual comparisons can be found in supplementary material. We will make the code publicly available for encouraging comparisons and improvements from the community.

\section{Conclusion}

In this work, we have proposed a deep network, named KinD, for low-light enhancement. Inspired by Retinex theory, the proposed network decomposes images into the reflectance and illumination layers. The decomposition consequently decouples the original space into two smaller subspaces. As ground-truth reflectance and illumination information is in short, the network is alternatively trained using paired images captured under different light/exposure conditions. To remove the degradations previously hidden in the darkness, the proposed KinD builds a restoration module.
A mapping function has also been learned in KinD, which better fits the actual situations than the traditional gamma correction, and flexibly adjusts light levels. Extensive experiments demonstrated the clear advantages of our design over the state-of-the-art alternatives. In the current version, KinD takes less than 50ms to handle an image in VGA resolution on a Nvidia 2080Ti GPU. By applying techniques like MobileNet or quantization, our KinD can be further accelerated. 

\bibliographystyle{ieeetr}
\bibliography{mybib}

\end{document}